\documentclass[letter]{ieice}

\usepackage[dvips]{graphicx}
\field{}

\title{Extending and Automating Basic Probability Theory with Propositional Computability Logic}
\authorlist{
\authorentry{Keehang Kwon}{m}{labelA}
}
\affiliate[labelA]{The author is a professor of  
Computer Eng., DongA University. email:khkwon@dau.ac.kr}

\received{2003}{1}{1}
\revised{2003}{1}{1}
\finalreceived{2003}{1}{1}

\newcommand{\sqc}{\mbox{\small \raisebox{0.0cm}{$\bigtriangleup$}}} 

\newcommand{\sqd}{\mbox{\small \raisebox{0.049cm}{$\bigtriangledown$}}} 

\newcommand{\gneg}{\mbox{\small $\neg$}}                  
\newcommand{\mli}{\hspace{2pt}\mbox{\small $\rightarrow$}\hspace{2pt}}                      
\newcommand{\mld}{\hspace{2pt}\mbox{\small $\vee$}\hspace{2pt}}     
\newcommand{\mlc}{\hspace{2pt}\mbox{\small $\wedge$}\hspace{2pt}}   
\newcommand{\add}{\hspace{2pt}\mbox{\small $\sqcup$}\hspace{2pt}}                     
\newcommand{\adc}{\hspace{2pt}\mbox{\small $\sqcap$}\hspace{2pt}} 

\newtheorem{theoremm}{Theorem}[section]
\newtheorem{factt}[theoremm]{Fact}
\newtheorem{corollaryy}[theoremm]{Corollary}
\newtheorem{definitionn}[theoremm]{Definition}
\newtheorem{thesiss}[theoremm]{Thesis}
\newtheorem{lemmaa}[theoremm]{Lemma}
\newtheorem{conventionn}[theoremm]{Convention}
\newtheorem{examplee}[theoremm]{Example}
\newtheorem{exercisee}[theoremm]{Exercise}
\newtheorem{remarkk}[theoremm]{Remark}

\newenvironment{theorem}{\begin{theoremm}}{\end{theoremm}}
\newenvironment{lemma}{\begin{lemmaa}}{\end{lemmaa}}

\newenvironment{numberedlist}
{\begin{list}{\makebox[20pt]{\hss(\arabic{itemno})\enspace}}
             {\usecounter{itemno}\labelwidth 20pt}}{\end{list}}

\newcounter{itemno}

\newcounter{itemno1}

\newcounter{itemno2}

\newcounter{exno}

\newcounter{defno}

\newcommand{\sep}{\;\vert\;}

\newcommand{\oprove}{\vdash\kern-.6em\lower.7ex\hbox{$\scriptstyle O$}\,}

\newcommand{\pderivation}{{\cal P}\kern -.1em\hbox{\rm -derivation}}
\newcommand{\pderivationl}{{\cal P}\kern -.1em\hbox{\em -derivation}}
\newcommand{\pderivable}{{\cal P}\kern -.1em\hbox{\rm -derivable}}
\newcommand{\pderivablel}{{\cal P}\kern -.1em\hbox{\em -derivable}}
\newcommand{\pderivations}{{\cal P}\kern -.1em\hbox{\rm -derivations}}
\newcommand{\pderivability}{{\cal P}\kern -.1em\hbox{\rm -derivability}}

\newsavebox{\lpartfig}
\newsavebox{\rpartfig}

\newenvironment{exmple}{
 \begingroup \begin{tabbing} \hspace{2em}\= \hspace{3em}\= \hspace{3em}\=
\hspace{3em}\= \hspace{3em}\= \hspace{3em}\= \kill}{
 \end{tabbing}\endgroup}

\newcommand{\mlip}{\hspace{2pt}\mbox{\small $\|$}\hspace{2pt}} 

\begin{document}
\maketitle
\begin{summary}
 Classical probability theory is formulated using sets. In this paper, we  extend classical  probability theory 
with propositional computability logic\cite{Jap03} (CoL).   Unlike other formalisms,
computability logic is built on the notion of events/games, which is central to probability theory.
 The probability theory based on CoL is therefore  useful for {\it automating} uncertainty reasoning.
We describe some basic properties of this new probability theory.
\end{summary}

\section{Introduction}\label{sec:intro}

Classical probability theory\cite{YG,Jay} is formulated using sets.
Unfortunately, the language of sets lacks expressiveness and is, in a sense, a low-level `assembly
language' of the probability theory.
In this paper, we  develop a `high-level approach' to classical  probability theory 
with propositional computability logic\cite{Jap03} (CoL).   
Unlike other formalisms such as sets, logic and linear logic,
computability logic is built on the notion of events/games, which is central to probability theory.
Therefore, CoL is a perfect place to begin the study of automating probability theory.

To be specific,  CoL is well-suited to describing complex (sequential/parallel) experiments and
events, and  more expressive than set operations.
In contrast, classical probability theory  -- based on $\cap,\cup$  --  is designed to   represent  mainly the  {\it simple/additive events} --  the events that occur
under a single experiment.

 Naturally, we need to talk about {\it composite/multiplicative events} -- events that  occur under two different experiments.
  Developing probability along this line requires a new, powerful language.
  For example, consider the following events $E_1, E_2$: \\
 
  $E_1$:  toss a coin two times  (we call the coin $1$)
  and get H,T in that order.
  
  $E_2$:  toss two dices (which we call $1,2$) and get at least one 5. \\
   
  \noindent Suppose a formalism  has the notion of $\sqc,\sqd$ (sequential-and/or)
  and $\mlc,\mld$ (parallel-and/or). Then $E_1$ would be written  as $H^1 \sqc T^1$. 
 Similarly,  $E_2$ would be written {\it concisely} as $(5^1 \mld 5^2)$.
 The formalism of classical probability theory fails to represent the above events in a concise way.

Computability logic\cite{Jap03} provides a formal and consistent way to 
represent  a wide class of  experiments and events. In particular, multiplicative experiments (sequential and parallel experiments) 
as well as additive experiments (choice-AND, choice-OR) can be represented in this formalism.

\section{The Event Formulas}\label{sec:logic}

The language is a variant of the propositional computability logic \cite{Jap03}. For simplicity, we
will not include  sequential operators. The class of
{\it event formulas} is described
by $E$-formulas given by the syntax rules below:
\begin{exmple}
  \>$E ::=$ \>   $  A \sep   \gneg E \sep E \add E \sep E \adc E \sep (E | E)
   $ \\  
\> \>   $  \sep  E \mlip  E  \sep  E \mlc E \sep E \mld E$ \\  
\end{exmple}
\noindent
In the above,  $A$    represents an  {\it atomic} event.  It is of the form $P(t_1,\ldots,t_n)$ where $P$ is either
a capital symbol or a number 
 and  each $t_i$ is a term.
For example, $D(6)$ represent the event where we get 6 from tossing a dice $D$.
For readability, we often write $6^D$ (or simply $6$) instead of $D(6)$.

Often events can be composed through the use of logical connectives. We will describe the definitions of these connectives.

First, we assume here experiments to be generated where the order is $not$ important.
Thus our {\it sample space} is  of the form

\[ \{ \vec{A}_1,\vec{A}_2,\ldots     \} \]

\noindent 
where  $\vec{A}$ represents an {\it unordered} tuple of   atomic events. 
Let $\vec{A}_i = (A^i_1,  \ldots, A^i_n)$.
As we shall see, the above set can  be rewritten as an event formula
in {\it set normal form} of the form

\[ (A^1_1 \mlc \ldots\mlc A^1_n) \add (A^2_1 \mlc \ldots\mlc A^2_n) \add \ldots \]

\noindent In the sequel, we use the above two forms interchangeably.

An {\it event space} is a subset of the sample space. In the
sequel,  we introduce a mapping $*$ which converts
an event formula $E$ to an event space $\{ \vec{A}_1,\ldots, \vec{A}_n \}$ of mutually exclusive points.  This mapping often makes it much easier to compute the probability of $E$.
 That is, $p(E) = p(E^*) = p(\vec{A}_1) + \ldots + p(\vec{A}_n)$.

As mentioned earlier,  $P(t_1,\ldots,t_n)$ represents an atomic event. In addition,

\[ P(t_1,\ldots,t_n)^* =  \{ P(t_1,\ldots,t_n) \} \]

The event $\gneg E$ represents  the complement of the event   $E$ relative to the universe $U$. 

\[ (\gneg E)^* = U - E^* \]

Suppose the events $E$ and $F$ are the outcomes of a single experiment.  In that case,
the choice-OR event $E \add F$ represents  the event in which only one of  the  event $E$  and event $F$  happen. For example, $4^D \add 5^D$ represents the event that  we get either  4 or  5 when a dice $D$ is   tossed.  This operation
corresponds to the set union operation. 

\[ (E \add F)^* = E^* \cup F^* \]

Suppose the events $E$ and $F$ are the outcomes of a single experiment.  In that case,
the choice-AND event $E \adc F$ represents  the event in which  both event $E$  and event $F$  happen. For example, $(2\add 4 \add 6) \adc (2 \add 3\add 5)$ represents the event that  we get both an even number 
and a prime number in a single coin toss.  This operation
corresponds to the set intersection operation.

\[ (E \adc F)^* = E^* \cap F^* \]

The additive conditional event  $E | F$ represents the dependency between $E$ and $F$:
the event in which  the event $E$  happens  given  $F$ has occurred under a single experiment.

 We can generalize the definition of events  to joint events to deal with multiple experiments.
 Here are some definitions.

The parallel-AND event $E \mlc F$ represents  the event in which both  $E$  and  $F$  occur  under two different experiments. For example, $(H^1 \mlc T^2) \add (H^2 \mlc T^1)$ represents the event that  we get one head and one tail when two coins are tossed.  It is defined by the following:

\[ (E \mlc F)^* = E^* \bar{\times} F^* \]

 Here, the (unordered, flattened) Cartesian conjunction of two sets $S$ and $T$, $S \bar{\times} T$ is the
 following:

\[ S \bar{\times} T = \{ (a_1,\ldots,a_m,b_1,\ldots,b_n) \} \]
\noindent where $(a_1,\ldots,a_m)\in S$ and  $(b_1,\ldots,b_n)\in T$.

\noindent For example, let $E = \{ (0,1), (1,2) \}$ and $F = \{ 0,1 \}$. Then
 $E \bar{\times} F = \{ (0,1,0),(0,1,1),(1,2,0),(1,2,1) \}$.

\ The parallel-OR event $E \mld F$ represents  the event in which at least one of    event $E$  and event $F$  happen under two different experiments. For example, $((4^1\add 5^1) \mld (4^2\add 5^2) $ represents the event that  we get at least one 4 or one 5 when two dices 1,2 are tossed. Formally,

 \[ (E \mld F) =_{def} (E \mlc F) \add (\gneg E \mlc F) \add (E \mlc \gneg F)\]

The parallel conditional event 
$E \mlip F$ (usually written as $F \mli E$ in logic)
represents  the event in which   the event $E$  happens  given  $F$ has occurred before/after
$E$ under two  experiments.

The following theorem extends  the traditional probability theory with
new symbols.  These  properties are  rather well-known.
Note that, sometimes, $p(E | F)$ is not available.
In that case, $p(E \adc F)$  can be computed using Rules (6)--(8).

\begin{theorem}\label{def:semantics}
Let $E$ be an event. Then the following properties hold.

\begin{numberedlist}



\item $p(\gneg E) = 1- p(E)$   \% complement of E

\item $p(E \add F) = p(E) + p(F) - p(E \adc F)$ \% choice-or

provided that $E$ and $F$ are the outcomes under a single experiment. Otherwise, $p(E \add F)$ is undetermined.
For example,     $H^C \add T^C$ represents the event that $H$ or $T$ occur  in a  single coin toss.
Now, it is easy to see that $p(H^C \add T^C) = 1$. However, $p(H^C \add T^D)$ is not determined.

\item $p(E \adc F) = p(E)p(F | E) = p(F)p(E | F)$ \% choice-and

provided that $E$ and $F$ are the outcomes under a single experiment. Otherwise, $p(E \adc F)$ is undetermined.
For example,  $H^C \adc T^C$ represents the event that $H$ and $T$ occur  in a  single coin toss. Now, $p(H \adc T) = 0$.
Note also that $p(H^C \adc H^C) = p(H^C)$.
 $p(E | F) $  is the conditional probability of F given E in a single experiment.
  For example,  $H^C |  T^C$ represents the event that  $H$ occurs given $T$ occurs in a  single coin toss. Now, 
  $p(H^C | T^C) = 0$.

\item  $p(E \mld F) = p(E \mlc F) + p(\gneg E \mlc F)  + p(E \mlc \gneg F) = $ \\
          $1 - p(\gneg E \mlc \gneg F)$  \% parallel-or 
 
 For example, $p(H \mld 6) = 1 - p(T \mlc (1\add 2\add 3\add 4\add 5) )= 1- 10/24 = 14/24$.
 
 

\item    $p(E \mlc F) = p(E)p(F \mlip E) = p(F)p(E \mlip F)$  \% parallel-and

For example, suppose two coins are tossed.    Now,   $p(H^1 \mlc H^2) = 1/4$. In addition, suppose a coin and a dice are tossed simultaneously.
Then $p(H^C \mlc 6^D) = p(H^C)p(6^D)= 1/12$.


   





\noindent         \% Below, $p$ computes the probability of  an event space rather than  an event formula.       
\item $p(E) = p(E^*)$ \% E* is the event space  of E.

For example,  $((3 \add 4) \adc 4)^*$ is $\{ 4 \}$.

\item $p(\{ \vec{A}_1,\ldots,\vec{A}_n \}) =  p(\{ \vec{A}_1 \}) + \ldots+ p(\{ \vec{A}_n \})$ \% event space with $n$ mutually exclusive elements.

\item $p(\{ (A_1,\ldots,A_n) \}) = p(A_1\mlc\ldots\mlc A_n)$ \%  event space with  single element. 
Here, each $A_i$ is an atomic event. 


         
\end{numberedlist}
\end{theorem}

\noindent In the above, we list some properties of our probability theory.
Following CoL, our language supports an additional kind of events: {\it predicate events}. Predicate events represent the events that have been completed
 such as ``2 is a prime number".  
 We use small letters $a,b,c,\ldots$ to denote predicate events, while we use    $A,B,C,3,5,\ldots$ to denote normal events.
 For example, $even(4), prime(2)$ are predicate events and $ 3,Dice(3)$ are  normal events.
 
 For this reason, we add the following rule:
 
\[ p(a \mlc a) =  p(a \mld a) = p(a). \]
\noindent 
  For example, let $a$ be a (uncertain) statement
that there are aliens in Vega. Then it is easy to see that these  rules
hold, as every occurrence of  $a$ represents the same event here.

\section{Examples }\label{sec:modules}

Let us consider the following event $E$  where 
$E$ = roll a dice and get 4 or 5.
The probabilities of $E$  and $E \mld E$ is  the following: \\

$p(E)$ =   $p(4 \add 5 )$ = $p(4) +  p(5)$  = $1/3$ \\

$p(E_1 \mld E_2) = 1 - (2/3 \times 2/3) = 5/9$. \\

\noindent As another example,  $(H^1 \mlc H^2) \add (H^1 \mlc T^2) \add (T^1\mlc H^2)$ represents the event that
at least one head comes up when two coins are tossed. 
Now, it is easy to see that  $p((H^1 \mlc H^2) \add (H^1 \mlc T^2) \add (T^1\mlc H^2))$ = 3/4.  \\

As the last example, suppose two dice 1,2 are tossed and we get 6 from  one dice. What is the probability that 
we get 5 from another dice?  This kind of problem is very cumbersome to represent/solve 
in classical probability
theory. Fortunately, it can  be represented/solved from the above formula in a concise way.
It is shown below: \\

 
\noindent  \%  Computing the following probability  requires  converting the event to its event space via the rules (6)--(8). \\
 
\noindent $\begin{array}{l}
  p(((6^1 \mlc 5^2)\add (6^2 \mlc 5^1)) \adc (6^1 \mld 6^2)) =  \\
     p(\{ (5^1,6^2),(6^1,5^2)\} \cap \{ (6^1,1^2),\ldots, \\ (6^1,6^2), (1^1,6^2),\ldots,(5^1,6^2) \}) =  \\
     p(\{ (5^1,6^2),(6^1,5^2) \}) =  p(5^1 \mlc 6^2) + p(6^1 \mlc 5^2) = 2/36   \\
\end{array}  $   \\ \\  \vspace{7pt}

\noindent \% Computing the following   does not require  converting the event to its event space. The rule (4) is used here. \\

\noindent  \( \begin{array}{l}
  p(6^1 \mld 6^2) = p(6^1 \mlc 6^2)+ p(\gneg 6^1 \mlc 6^2) + p( 6^1 \mlc \gneg 6^2) = \\
   11/36 \\
    \end{array}  \)    \\ \\ \vspace{7pt}
  
  From these two, we obtain the following (via Rule (3)) in a purely  algorithmic way: \\ \\
  
\noindent  $\begin{array}{l}
   p(((6^1 \mlc 5^2)\add (6^2 \mlc 5^1)) | (6^1 \mld 6^2)) = 2/11 \\
   \end{array}$  \ \   \vspace{7pt}


\section{Two Versions of the Bayes’ Rule}\label{sec:bayes}

In this section, we raise questions related to the interpretation of
$\cap$ in the Bayes rule. Most textbooks  interpret $\cap$ in an ambiguous, 
confusing way: sometimes as $\adc$, and  as $\mlc$  in others.
This is problematic, especially in automating probabilistic inference.

In considering automation of probability, it is very
problematic/cumbersome to use
the Bayes rule in its current form. Suppose
$E$ can be partitioned into $k$ disjoint events $E_1,\ldots,E_k$.
Understanding $\cap$ as $\adc$, the Bayes rule can be written as:

\[ p(E_i | F) = \frac{p(E_i \adc F)}{p(E_1 \adc F)+\ldots+p(E_k \adc F)} \]

or

\[ p(E_i | F) = \frac{p(E_i \adc F)}{p(E_1)p(F | E_1)+\ldots+p(E_k)p(F | E_k)} \]

\noindent This rule can easily be generalized to $\mlc$ as follows:

\[ p(E_i \mlip F ) = \frac{p(E_i \mlc F)}{p(E_1\mlc F)+\ldots+p(E_k\mlc F)}\]

or

\[ p(E_i \mlip F) = \frac{p(E_i \mlc F)}{p(E_1)p(F \mlip E_1)+\ldots+p(E_k)
  p(F \mlip E_k)} \]

That is, we need {\it two} versions of the Bayes rule and it is crucial to apply the correct version
to get the correct answer.
As a well-known example of the Bayes rule, consider the problem of sending 0 or 1 over a noisy channel.
Let $R(0)$ be the event that a 0 is received.
Let $T(0)$ be the event that a 0 is transmitted.
Let $R(1)$ be the event that a 1 is received.
Let $T(1)$ be the event that a 1 is transmitted.
Now the question is: what is the probability of $0$ having been transmitted, given 0 is received?
In solving this kind of problem, 
it is required to use Bayes rule on  $\mlc$, rather than on
$\adc$. This is so because $p(T(0) \cap R(0)) = p(T(0) \adc R(0)) = 0$.

\section{Conclusion}\label{sec:conc}

Computability logic \cite{Jap03}--\cite{JapCL12} provides a rich vocaburary to 
represent  a wide class of  experiments and events such  as sequential/toggling operations.  For this reason, we believe that probability theory based on computability logic will prove useful
for uncertainty reasoning in the future.

\bibliographystyle{ieicetr}



\end{document}